\documentclass[runningheads]{llncs}

\usepackage{eccv}

\usepackage{eccvabbrv}

\usepackage{graphicx}
\usepackage{booktabs}

\usepackage[accsupp]{axessibility}  

\usepackage{hyperref}

\usepackage{orcidlink}

\usepackage{makecell}
\usepackage{threeparttable}
\usepackage{comment}
\usepackage{subcaption}

\newcommand{\alist}[1]{\begin{itemize}#1\end{itemize}}

\begin{document}

\title{Toward Trustworthy Robot-Assisted Sliding Palpation for Shallow Vessel Localisation with a Calibrated Digital Twin}

\titlerunning{Robot-Assisted Sliding Palpation for Shallow Vessel Localisation}

\author{%
Piotr Blaszyk\orcidlink{0009-0004-4427-4908} \and
Wen Fan\orcidlink{0009-0009-4239-1242} \and
Kaizhong Deng\orcidlink{0009-0007-4206-5282} \and
Daniel Elson\orcidlink{0000-0002-5578-3941} \and
Dandan Zhang\orcidlink{0000-0001-7649-7605}%
}

\authorrunning{P. Blaszyk et al.}

\institute{Imperial College London, UK}

\maketitle

\begin{abstract}
Reliable localisation of shallow subsurface vessels matters for safe robot-assisted venous access and vessel-aware manipulation, but collecting diverse tactile data on physical hardware is costly, slow, and can degrade soft vision-based tactile sensors. We present a robot-assisted sliding-palpation framework in which a calibrated digital twin generates labelled tactile sequences, reducing dependence on real-world data collection. The twin models sensor–vessel contact, is calibrated against real palpation trajectories by Bayesian-optimisation-based domain adaptation, and is randomised over sliding direction and contact conditions. A spatio-temporal graph neural network trained on simulated marker trajectories performs per-node vessel classification and yields a human-verifiable top-view localisation map through 2D-to-3D-to-2D geometric projection. Using three datasets (Sim, Silicone, and Meat, the latter a raw-meat phantom with vessel models at nominal depths of 0-30\,mm) we evaluate four [train]$\rightarrow$[test] models: Sim$\rightarrow$Sim, Sim$\rightarrow$Silicone, Sim$\rightarrow$Meat, and Meat$\rightarrow$Silicone. The calibrated twin, validated on four canonical interactions, yields a simulated-to-real marker-alignment MAE of 0.50 mm at the point of deepest contact. After reprojection onto a 1 mm top-view grid, with each model's decision threshold set to favour few false alarms over high sensitivity, predicted vessel pixels lie on average 1.05-5.49 mm from the nearest true vessel pixel across the four models (1.05-1.31 mm for all but Sim$\rightarrow$Meat, whose larger domain shift marks the current limit of simulation transfer). These results demonstrate progress toward trustworthy tactile palpation through calibrated simulation, interpretable localisation, and transparent reporting of cross-domain degradation. Code, model weights, and data are publicly available on GitHub and Zenodo.
\footnotemark[1]
    
\end{abstract}

\footnotetext[1]{We give Zenodo DOIs here, as they are permanent.
Simulation and learning code~\cite{blaszyk_shallow-vessel-palpation-simulator-and-ai_2026}:
\url{https://doi.org/10.5281/zenodo.21958186}.
Robot control code~\cite{blaszyk_shallow-vessel-palpation-robot-control_2026}:
\url{https://doi.org/10.5281/zenodo.21958190}.
Datasets and model weights~\cite{blaszyk_shallow-vessel-palpation-dataset_2026}:
\url{https://doi.org/10.5281/zenodo.21958107}.}

\section{Introduction}

\begin{figure}
    \centering
    \includegraphics[width=0.900\linewidth]{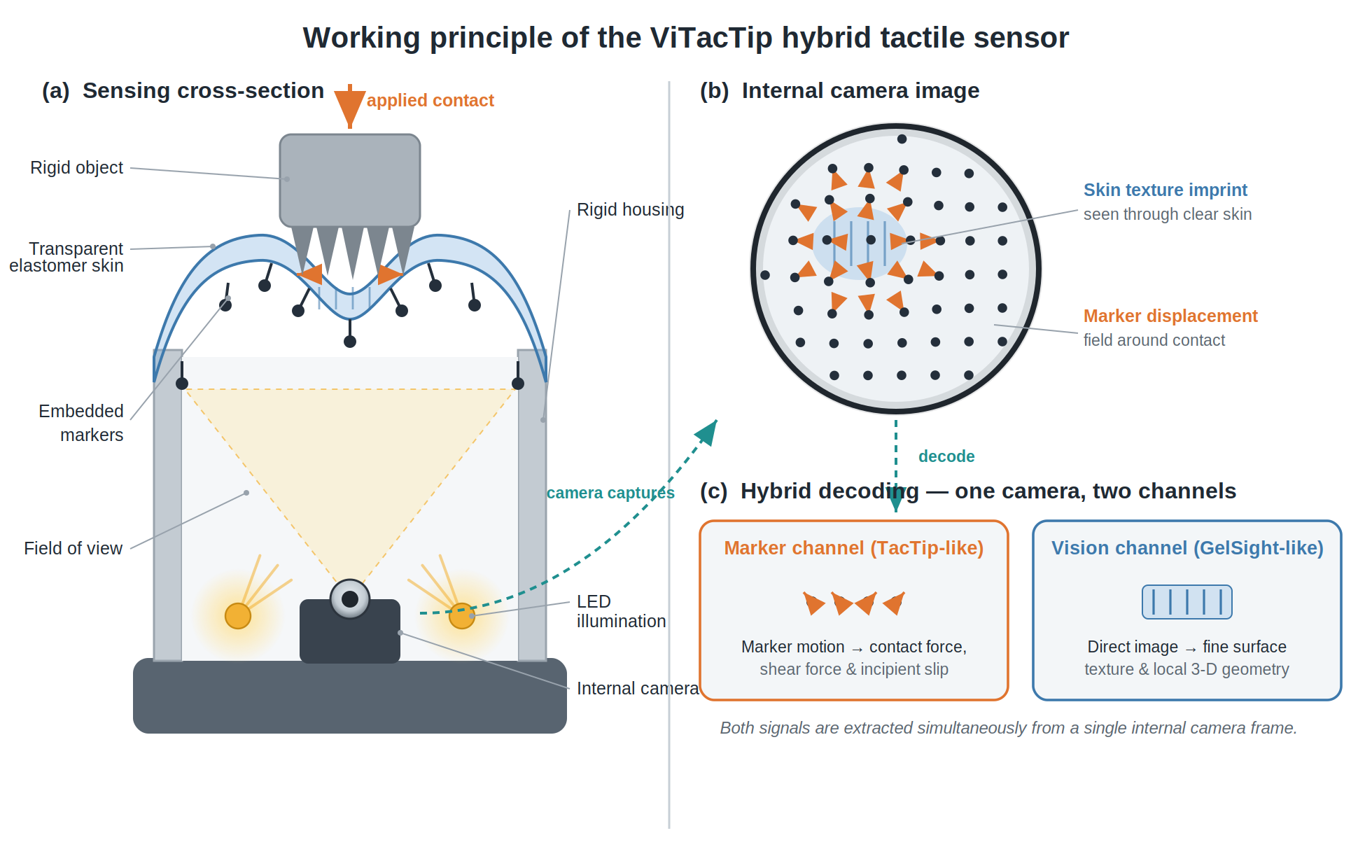}
    \caption{The working principle of the ViTacTip sensor used in this work. This figure was inspired by the work of Fan et al.~\cite{81}.}
    \label{fig:vitactip_working_principle}
\end{figure}

Shallow blood vessel detection is an important capability in robot-assisted minimally invasive surgery (RAMIS), where it helps prevent accidental vessel injury during manipulation and tissue interaction~\cite{Penza2017}. It is also essential for clinical procedures such as catheterisation for intravenous injections~\cite{10160848}, blood sampling, and interventional treatments including cardiac stenting~\cite{MoussaPacha2018}. In oncology applications, the spatial distribution of vasculature surrounding a tumour is a key factor when determining electrode placement for irreversible electroporation~\cite{Kos2015}.

Vessel detection is supported by several imaging modalities. Magnetic resonance imaging (MRI) and computed tomography (CT) can map vessels at high resolution and depth but are expensive and, for CT, expose the patient to radiation. Ultrasound (US) is the clinical gold standard for intra-operative detection of shallow and deep vessels~\cite{83}; near-infrared imaging (NIR) is common in phlebotomy for superficial vessels~\cite{vd-nir}; and optical coherence elastography (OCE) is an emerging approach with finer resolution than either~\cite{vd-oe}. All rely on specialised imaging equipment.

Their cost is a key limitation: NIR and US probes cost a few thousand dollars (e.g., \$1,800~\cite{nir-cost} and \$4,800~\cite{us-cost}), and commercial ophthalmic optical coherence tomography (OCT) probes were reported at \$40,000--\$150,000 in 2021~\cite{oct_cost}. They also need specialised maintenance, infrastructure, and trained operators, which hinders deployment in resource-constrained settings, particularly in low- and middle-income countries~\cite{frugal_innovation}, motivating lower-cost approaches to shallow vessel detection.

\begin{figure}[t]
    \centering
    \includegraphics[width=\linewidth]{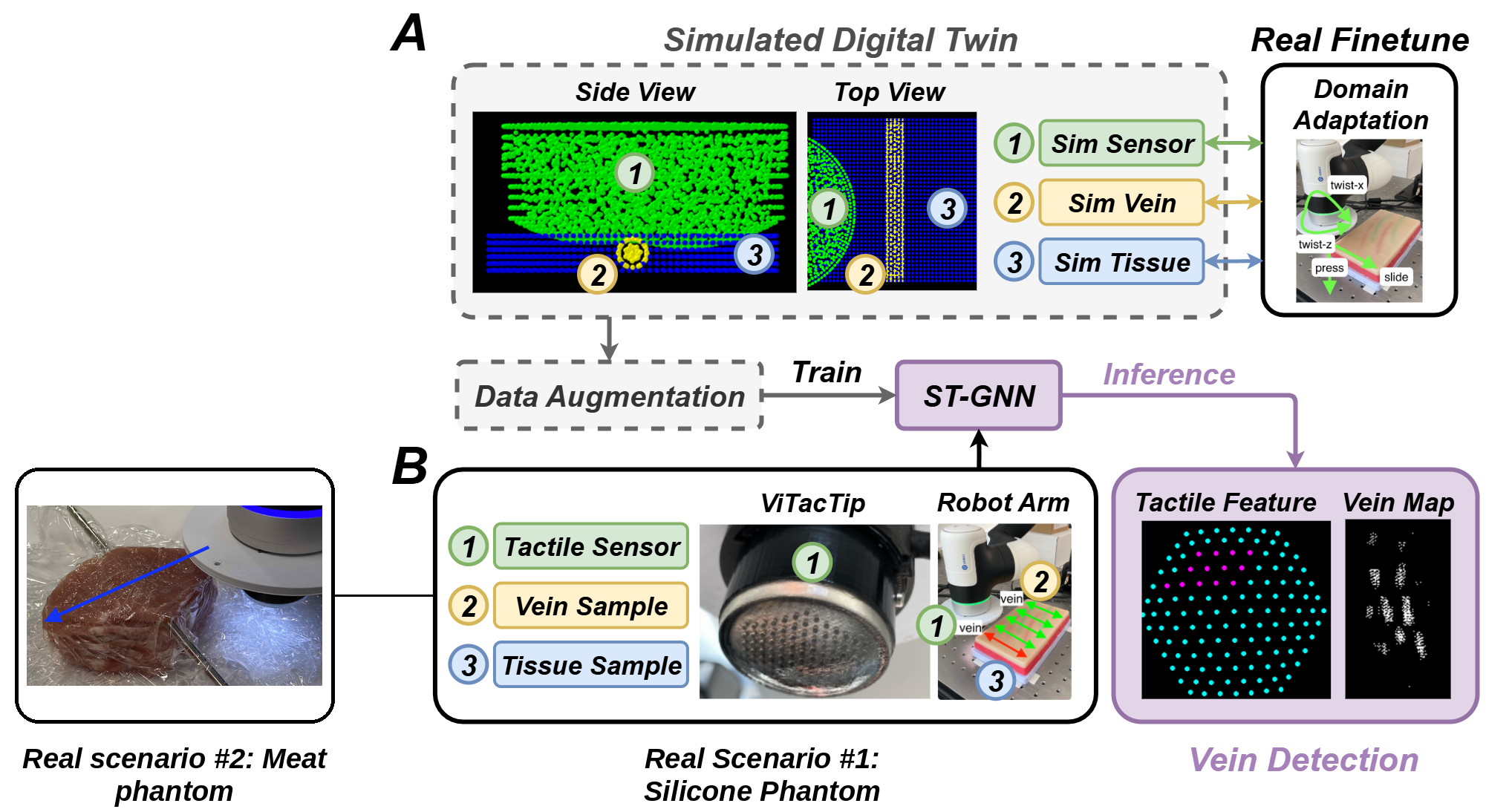}
    \caption{Vessel-detection pipeline. Domain adaptation calibrates the sensor Young's modulus and sensor–vessel contact stiffness, while domain randomisation varies the sliding direction and sensor–vessel contact parameters. Training data are generated using the DiffTactile-based digital twin, and an ST-GNN performs per-node binary classification on the marker graph (magenta: vessel; cyan: no vessel). Robot kinematics are then used to project these predictions into a top-view vessel map (white: predicted vessel). Real ViTacTip trajectories collected on the silicone vascular phantom (Real Scenario \#1) and the raw-meat phantom (Real Scenario \#2) provide labelled tactile videos for evaluating the ST-GNN.}
    \vspace{-0.3cm}
    \label{fig:holistic-sys-arch}
\end{figure}


Vision-based tactile sensors (VBTSs) provide a compact and comparatively low-cost approach to high-resolution tactile perception. By imaging the deformation of a compliant sensing surface during contact, VBTSs can capture rich information about local geometry and contact mechanics~\cite{gelsight,digit,abad2020visuotactile}. Fig.~\ref{fig:vitactip_working_principle} illustrates the operating principle of one such sensor, the ViTacTip~\cite{81}. A key practical advantage of VBTSs is their affordability. Commercial devices such as the GelSight Mini and DIGIT cost approximately \$500 and \$350, respectively~\cite{gelsight-mini-cost,digit-cost}, while laboratory-built designs can reduce fabrication costs further. For example, MagicTac reports a minimum manufacturing cost of only £4.76~\cite{fan2024magictac}. In comparison, point-of-care ultrasound systems typically cost several thousand pounds, and conventional cart-based systems can cost tens of thousands of pounds~\cite{patlan2016automatic}. Ultrasound remains an established modality for vascular access and offers direct imaging of subsurface anatomy at substantially greater depths~\cite{lamperti2012ultrasound}. However, for the localisation of shallow subsurface vessels, the substantially lower hardware cost, compact form factor, and ease of robotic integration of VBTSs motivate their investigation as a complementary sensing modality.

Beyond their cost advantage, VBTSs provide a high degree of sensing versatility. The rich visual information generated by contact-induced deformation can be used to infer a range of tactile quantities, including surface geometry, contact location, force, and slip~\cite{fan2025crystaltac}. Recent developments further demonstrate the potential to integrate multiple sensing modalities within a single compact robotic end-effector. For example, MagicGripper combines tactile, visual, and proximity sensing to support diverse contact-rich manipulation tasks~\cite{fan2025magicgripper}. This multifunctionality enables a common sensing platform to support different perception and manipulation objectives through task-specific processing and inference, without requiring dedicated hardware for each function. Such flexibility is particularly valuable for compact and resource-constrained robotic systems and further motivates the use of VBTSs for robot-assisted palpation.

In the intended clinical workflow, an end-effector-mounted VBTS performs a brief tactile pre-scan of the skin to localise superficial vessels before any invasive action. The resulting vessel map can support automated venous access tasks, such as intravenous cannulation and blood sampling, or vessel-aware manipulation during RAMIS. By presenting the localisation result as a human-verifiable top-view map, the framework retains clinician oversight and enables confirmation of vessel location before needle insertion or instrument interaction.



Vessel detection from VBTS deformation patterns typically needs learning-based models, and hence a large and diverse dataset spanning the contact conditions met during palpation. Collecting it on physical hardware is slow and labour-intensive, and the soft elastomer and delicate optics can degrade under repeated interaction. Digital twins (DTs) address this by simulating tactile interactions at scale without physical wear~\cite{fan2023digital}, generating common and rare contact scenarios with automatic ground-truth labels and greater diversity, which can improve generalisation over limited real datasets~\cite{Mazumder2023}. Two complementary strategies mitigate the sim-to-real gap: domain adaptation, which aligns DT parameters with their real-world counterparts, and domain randomisation, which adds variability to approximate real conditions~\cite{sim2real}.

To the best of our knowledge, this study is the first to combine a VBTS with sliding palpation and a DT for shallow vessel detection, and asks how accurately and reliably shallow vessels can be localised from VBTS deformation observed while sliding across them.
The \textbf{main contributions} of this work are as follows:
\alist{
\item We formulate shallow-vessel localisation from continuous robot-executed sliding palpation using a vision-based tactile sensor, enabling spatially continuous sensing along the scanning trajectory.

\item We develop and calibrate a digital twin of sensor--vessel interaction and use domain-randomised simulation to train a spatio-temporal graph neural network (ST-GNN) for per-node vessel segmentation without silicone-phantom training data.

\item We evaluate sim-to-real transfer across two physical experimental domains: a silicone vascular phantom and a raw-meat phantom containing tubular vessel surrogates at nominal depths of \(0\)--\(30\,\mathrm{mm}\), including explicit analysis of performance degradation under the larger domain shift, seed-to-seed variability, and a temporal-window ablation.

\item We transform tactile predictions using robot kinematics into a human-verifiable top-view vessel map, evaluated per pixel on all three datasets, that can provide spatial input for subsequent robot planning.
}

\section{Related work}\label{sec:related-work}

\begin{table*}[t]
    \centering
    \begin{threeparttable}
    \caption{Comparison of related work: part 1}
    \label{tab:comparison-1}
    \begin{tabular}{|l|l|l|l|l|l|l|l|}
    \hline
        \textbf{Project} & \textbf{Target} & \textbf{Sensor} & \textbf{Palpation} & \textbf{Approach} & \textbf{BP} & \textbf{Train} & \textbf{Test} \\ \hline
        \textbf{MiniTac} \cite{84} & T & VBTS & T & M & n/a & R & P, EV \\ \hline
        \textbf{DIGIT Pinki} \cite{80} & T & VBTS & T & M & n/a & R & P, EV \\ \hline
        \textbf{Bewley et al.} \cite{74} & T & VBTS & T & EE & n/a & R & P \\ \hline
        \textbf{Beasley et al.} \cite{Beasley} & BV & Capacitive & S & EE & Y & n/a & P, IV \\ \hline
        \textbf{Hampson et al.} \cite{Hampson2023} & BV & Tactile array & T & EE & Y & n/a & P \\ \hline
        \textbf{3D CNN} \cite{Chen2022} & BV & NIR & none & EE & N & R & IV \\ \hline
        \textbf{Ours} & BV & VBTS & S & EE & N & DT & P, M \\ \hline
    \end{tabular}
    \begin{tablenotes}[flushleft]
        \footnotesize
        \item \textbf{Abbreviations:} target (T = tumour, BV = blood vessel), palpation (T = tapping, S = sliding), approach (M = modular, EE = end-to-end), BP = use blood pulsation data (N = no, Y = yes), train set (R = real-world data, DT = digital twin), test set (P = synthetic phantom, EV = ex-vivo human tissue, IV = in-vivo human tissue, M = raw meat phantom)
    \end{tablenotes}
        \vspace{-0.4cm}
    \end{threeparttable}
\end{table*}
\begin{table}[t]
    \centering
    \footnotesize
    \setlength{\tabcolsep}{4pt}
    \begin{threeparttable}
    \caption{Comparison of related work: part 2}
    \label{tab:comparison-2}
    \begin{tabular}{|l|l|l|l|}
    \hline
        \textbf{Project} & \textbf{Model} & \textbf{Accuracy [\%]} & \textbf{Error [mm]} \\ \hline
        \textbf{MiniTac} \cite{84} & MLP, SVM & 100 & n/a \\ \hline
        \textbf{DIGIT Pinki} \cite{80} & RN18, VMAE & 98, 100 & n/a \\ \hline
        \textbf{Bewley et al.} \cite{74} & LR, GKDE & 95, 70 & 1 \\ \hline
        \textbf{Beasley et al.} \cite{Beasley} & MMD, THR, LR & n/a & 1.3 \\ \hline
        \textbf{Hampson et al.} \cite{Hampson2023} & MM, GC & n/a & 0.2, 0.6, 6.0 \\ \hline
        \textbf{3D CNN} \cite{Chen2022} & 3D CNN & 87 & n/a \\ \hline
        \textbf{Ours}\tnote{a} & ST-GNN & n/a & 1.1, 1.2, 5.5, 1.3 \\ \hline
    \end{tabular}
    \begin{tablenotes}[flushleft]
        \footnotesize
        \item \textbf{Abbreviations:} RN18 = ResNet-18; VMAE = VideoMAE; LR = linear regression; GKDE = Gaussian kernel density estimation; MMD = min-max detection; THR = thresholding; MM = mathematical model; GC = Gaussian curve; ST-GNN = spatio-temporal graph neural network.
        \item[a] Mean L2 error of the top-view vessel map (Table~\ref{tab:localisation-map}), for Sim$\rightarrow$Sim, Sim$\rightarrow$Silicone, Sim$\rightarrow$Meat, Meat$\rightarrow$Silicone in that order.
    \end{tablenotes}
    \end{threeparttable}
    \vspace{-0.5cm}
\end{table}

\subsection{Vision-based Tactile Sensors for Tactile Palpation}

Tables~\ref{tab:comparison-1} and~\ref{tab:comparison-2} summarise related studies on VBTS-based tumour palpation (whose stiffness-variation principle mirrors vessel localisation) and on vessel detection with other modalities. The `error' column is the mean error for related works and, for ours, the mean distance from each predicted vessel pixel to the nearest true pixel of the top-view map. Beasley et al.~\cite{Beasley} also slide, but stop for 4\,s at each collection point; the remaining works tap. MiniTac~\cite{84} and DIGIT Pinki~\cite{80} are modular (map sensor image to deformation, then classify), whereas Bewley et al.~\cite{74} classify end-to-end on a Voronoi graph of the markers.

\subsection{Digital Twin-based Data Augmentation}
Digital twins (DTs), high-fidelity virtual replicas in which sensing, dynamics, and interaction are simulated, have become an effective source of large-scale training data for robotic perception~\cite{Yao2023}. For tactile sensing they yield large, diverse contact datasets without the cost and hardware wear of physical experiments; combining physics-based simulation with data-driven models reproduces contact deformation and sensor noise well enough to approximate physical tactile sensors~\cite{physics-inspired-digital-twins} and to label data across many interaction conditions~\cite{Mazumder2023}. DT-based generation is commonly paired with sim-to-real techniques~\cite{sim2real}: domain adaptation aligns simulated and real sensor characteristics, while domain randomisation adds controlled variability in textures, lighting, dynamics, and sensor parameters~\cite{sizhe2025}.

\section{Methods}

\subsection{Justification of Choices}

\subsubsection{Choice of VBTS for Vessel Detection}
We select the ViTacTip~\cite{81} for its compact combination of vision and tactile sensing: a transparent elastomeric ``see-through'' skin over a biomimetic marker structure lets the internal camera observe surface features while the markers track local strain and shear during sliding. Its reported low pose/force error and strong fine-texture discrimination~\cite{81} suit sliding palpation, where local deformation and shear reveal shallow-vessel stiffness variations, and the visual channel aids ground-truth identification of vessel locations.

\subsubsection{Choice of GNNs for Tactile Perception}

For marker-based VBTSs, graph neural networks (GNNs) have proven effective for tactile perception~\cite{82}: the markers form a grid whose displacements encode surface deformation, mapping naturally to a graph with markers as nodes and spatial neighbourhoods as edges. Relative to CNNs, GNNs model inter-marker dependencies more explicitly and perform better on marker-based tactile tasks~\cite{82}.

\subsection{System Overview}

Fig.~\ref{fig:holistic-sys-arch} illustrates the pipeline. We build a DT on DiffTactile~\cite{DiffTactile} comprising a rigid phantom with a single rigid vessel inclusion and a soft-body (FEM) VBTS, with a camera model mapping 3D marker positions to a synthetic marker image. Domain adaptation fits the sensor's Young's modulus and the sensor–vessel contact stiffness by Bayesian optimisation on a real sliding trajectory (Sec.~\ref{sec:da-methods}); the fitted twin is validated against four real trajectories (press, slide, twist-x, twist-z) by the marker-position error at deepest contact.

\begin{figure}[tbp]
    \centering
    \begin{subfigure}{0.48\textwidth}
        \centering
        \includegraphics[width=0.9\textwidth]{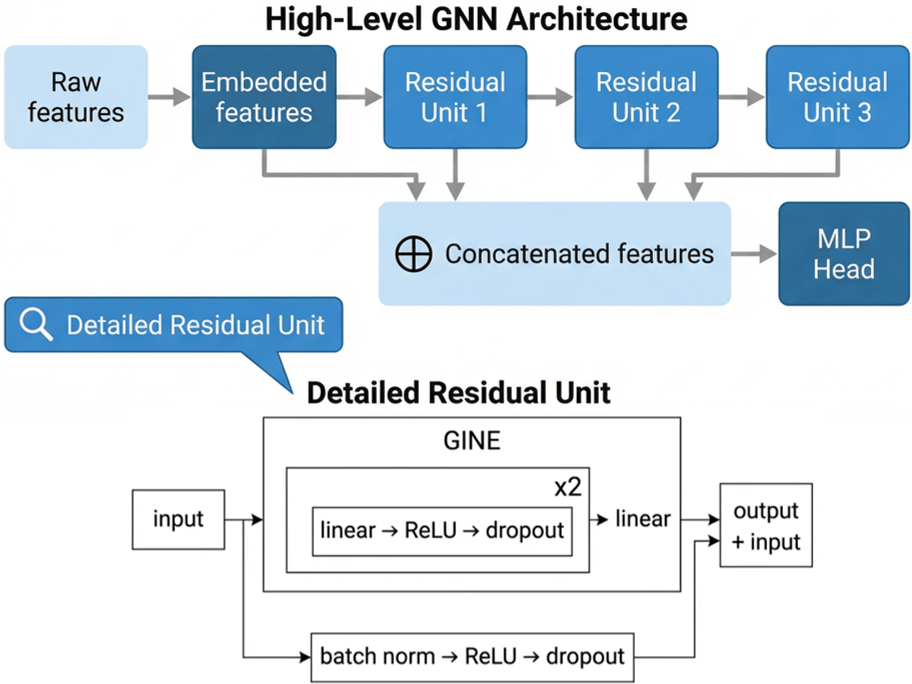}
        \caption{}
        \label{fig:gnn-arch}
    \end{subfigure}
    \hfill
    \begin{subfigure}{0.48\textwidth}
        \centering
        \includegraphics[width=0.9\textwidth]{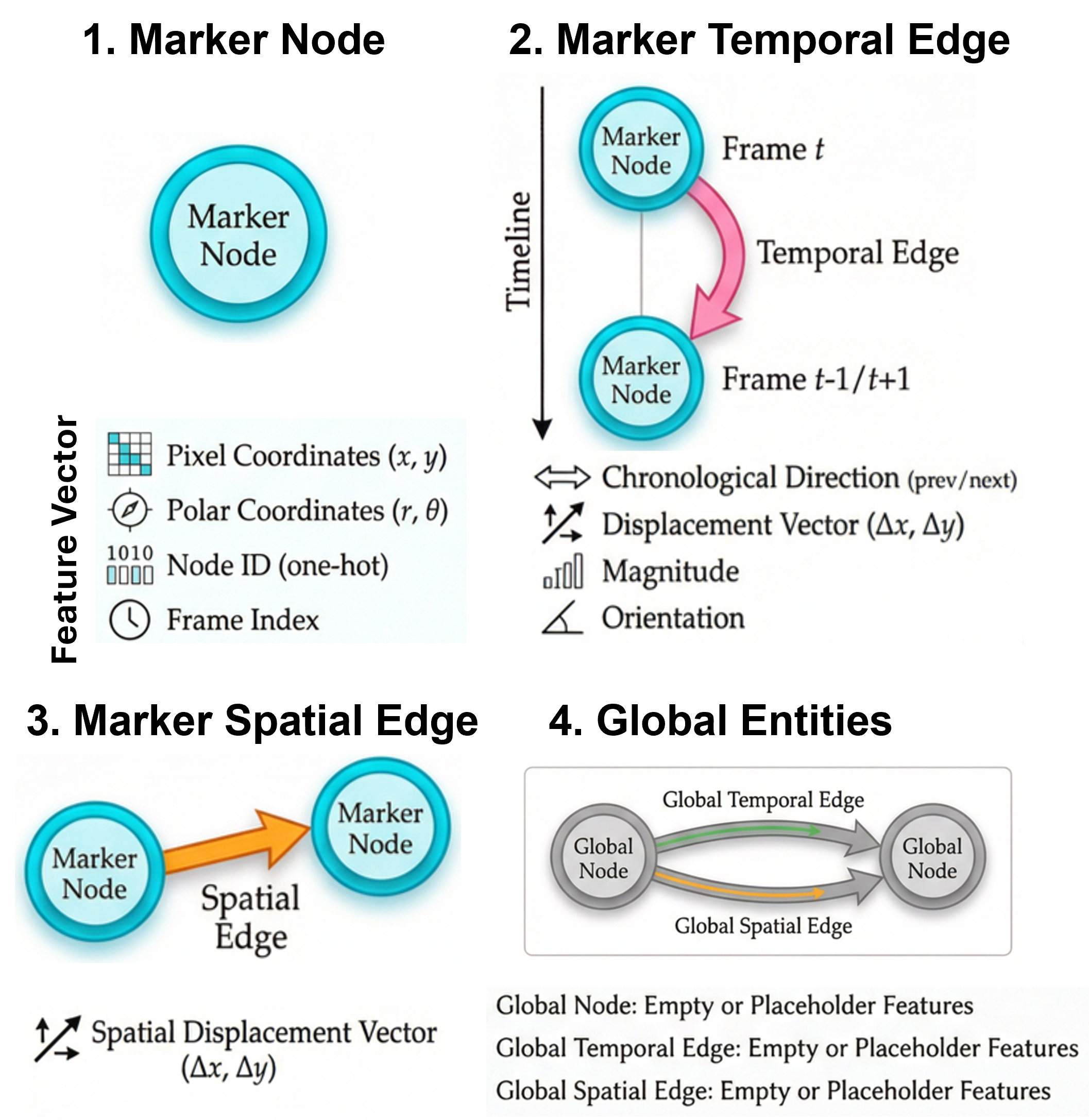}
        \caption{}
        \label{fig:graph-entity-types}
    \end{subfigure}
    \caption{(a) ST-GNN architecture. Raw node features (marker, global) and edge features (spatial, temporal, global-spatial, global-temporal) are projected to a shared embedding space and processed by 3 GINE (graph isomorphism network with edge support) convolution layers, each in a residual unit with dropout and batch normalisation, giving a 3-hop receptive field. The input and all 3 layer representations are concatenated by skip connections (late fusion) and an MLP head outputs one classification value per marker node. (b) Graph entity types and raw input features. Each frame has marker nodes and one global node; within a frame markers connect via spatial edges and to the global node via global-spatial edges; across adjacent frames each marker connects to itself via temporal edges and the global nodes via global-temporal edges.}
    \label{fig:gnn-combined}

    \vspace{-0.3cm}
\end{figure}

We then generate 500 simulated sliding trajectories (250 with and 250 without the vessel), the Sim dataset of Sec.~\ref{sec:dataset-definitions}, to train a deep spatio-temporal GNN (ST-GNN) for binary per-node vessel classification on the marker graph (Fig.~\ref{fig:gnn-combined})~\cite{sahili2023spatio}. Each sample is a 5-frame clip (window length chosen by the ablation in Table~\ref{tab:clip-len}); training applies deep supervision on all markers of all five frames, testing uses only the central frame. Every model is trained under five random seeds; the precision--recall curves report mean $\pm$ standard deviation, while the tabulated statistics and the top-view maps use the instance with the highest average precision on its test set. The trained ST-GNN is evaluated on Silicone (ten annotated videos of straight-line sliding across a vessel), from which we also produce a top-view vessel map, and on Meat.

\begin{table}[tb]
    \centering
    \footnotesize
    \caption{Temporal-window ablation: Sim$\rightarrow$Silicone trained with a clip of $L$ frames. Arrows give the desirable direction; \textbf{bold} marks the best value per column and \underline{underline} the runner-up. $L=5$ is used throughout the rest of this manuscript}
    \label{tab:clip-len}
    \begin{tabular}{lcc}
    \toprule
    $L$ & \textbf{FG IoU}$\uparrow$ & \textbf{AP}$\uparrow$ \\
    \midrule
    1 & $0.114 \pm 0.002$ & $0.209 \pm 0.005$ \\
    3 & $0.197 \pm 0.006$ & $0.201 \pm 0.001$ \\
    5 & $\mathbf{0.238 \pm 0.001}$ & $\mathbf{0.324 \pm 0.001}$ \\
    7 & \underline{$0.214 \pm 0.029$} & \underline{$0.311 \pm 0.006$} \\
    \bottomrule
    \end{tabular}
    \par\vspace{2pt}
    \begin{flushleft}\footnotesize
    $L$ = clip length in frames; FG IoU = foreground (vessel-present) IoU at threshold 0.5; AP = average precision (chance level 0.11). Mean $\pm$ standard deviation over three training seeds, on the Silicone test set.
    \end{flushleft}
    \vspace{-0.3cm}
\end{table}

\subsection{Digital Twin Construction}

The digital twin is a physics simulation, built on the Taichi-based DiffTactile~\cite{DiffTactile} soft-body simulator and its VBTS interfaces, that replicates sensor–vessel contact and renders synthetic marker positions matching the real sensor, yielding automatically labelled palpation sequences. Only the sensor is deformable: the \textit{ViTacTip} (Fig.~\ref{fig:holistic-sys-arch}) is modelled with the finite element method (FEM) from a simplified \texttt{gmsh} CAD model, with one homogeneous material replacing the real sensor's two and the 127 biomimetic tips omitted to reduce meshing and simulation cost. The vessel and the phantom are rigid bodies, the vessel static beneath the phantom surface. Only the sensor–vessel contact pair is enabled; the sensor–phantom pair is disabled, so the phantom's particles are present for visualisation only and do not affect the simulation. Contact is a penalty-based spring-damper formulation driven by signed distance functions (SDFs); in the sensor–vessel pair only the normal stiffness $k_n$ and normal damping $c_n$ are non-zero, while the tangential stiffness $k_t$ and Coulomb friction coefficient $\mu$ are zero for simplicity, so the vessel pushes on the sensor but never drags it. This works because the inertia of the deformable dome, combined with its finite elastic coupling to the rigid sensor base, shows up as bending during sensor acceleration, producing marker displacements that align well with real data despite the disabled sensor–phantom pair. A fisheye camera model projects simulated 3D markers to image space.

\subsection{Domain Adaptation and Randomisation}\label{sec:da-methods}

Domain adaptation fits two parameters by Bayesian optimisation (BO) over the black-box simulator: the sensor's Young's modulus $E$ and the sensor–vessel normal stiffness $k_n$ (Poisson's ratio, $c_n$, $k_t$ and $\mu$ are fixed). Each BO iteration replays two simulated slides: a vessel-absent one, scored by the mean absolute error (MAE) between simulated and real marker positions at deepest contact of a real slide on the silicone phantom, normalised by three inter-marker spacings, and a vessel-present one, rewarded for sufficient sensor deformation on vessel contact (the sensor's height above the vessel relative to the vessel-free run). The objective is their difference, searched by five random draws and five acquisition steps. The fitted configuration is validated, without further fitting, against four canonical real trajectories (press, twist-$z$, twist-$x$, slide) by the same MAE. Domain randomisation varies, per simulated trajectory, the sliding direction ($\pm 15^\circ$ about the nominal crossing direction) and the sensor–vessel $k_n \in [5\times10^{3}, 5\times10^{4}]$\,N/m and $c_n \in [0, 100]$\,N\,s/m; this $k_n$ range predates the final calibration and, by oversight, was not re-centred on the fitted value (Sec.~\ref{sec:da-results}) for want of another collection run, though the Sim data still look realistic on inspection.

\section{Experiments}

\subsection{Dataset Definitions}\label{sec:dataset-definitions}

We use three datasets, referred to throughout by these names: \textbf{Sim},
generated in the digital twin; \textbf{Silicone}, collected on the simpler
silicone phantom with shallow vessels; and \textbf{Meat}, collected on the more
challenging meat phantom with vessels at varying depths. From these we train
and evaluate four models, denoted [train dataset]$\rightarrow$[test dataset]:
\textbf{Sim$\rightarrow$Sim}, an in-domain model tested on held-out simulated
data; \textbf{Sim$\rightarrow$Silicone} and \textbf{Sim$\rightarrow$Meat}, which
measure sim-to-real transfer to each physical phantom; and
\textbf{Meat$\rightarrow$Silicone}, a real-data baseline against
Sim$\rightarrow$Silicone. The Sim$\rightarrow$* models share the same training
recipe and data and differ only in the test set.

\subsection{Common Phantom Designs}
Tactile localisation phantoms replicate local stiffness variations. Studies of tumour palpation typically embed a steel ball bearing in a silicone phantom~\cite{84}, varying inclusion depth (1--7\,mm)~\cite{84}, diameter (2--12\,mm)~\cite{80}, and tapping force (1--8\,N)~\cite{74}, sometimes validating on ex-vivo tissue with healthy and cancerous regions~\cite{84}. Vessel detection instead uses phantoms mimicking compliant vessels in soft tissue, e.g., silicone with heat-shrink tubing~\cite{Beasley} or latex tubing in foam~\cite{Hampson2023}, with occasional in-vivo validation on superficial vessels such as the temporal, carotid, or radial arteries~\cite{Chen2022} or the dorsalis pedis artery~\cite{Hou2025}.


None of the analysed works uses a raw-meat phantom. Meat's 10-layer stacked design allows the inclusion depth and type to be varied without fabricating multiple phantoms. Compared with the silicone phantom, the layered raw-pork substrate provides a more mechanically complex test domain and exhibited qualitatively lower sliding resistance during our experiments.

\subsection{Common Evaluation Metrics}

VBTS tumour detection usually forms a sensor deformation map and trains a classifier on it~\cite{84}, most commonly tumour vs.\ no tumour~\cite{84,80,74}; Bewley et al.~\cite{74} also classify tumour diameter (2--10\,mm) and depth (1--5\,mm) under varying normal force (1--8\,N). Vessel detection reports $n$-pixel accuracy (fraction of trials with prediction–truth distance $\leq n$ pixels)~\cite{Chen2022}, absolute $x$-axis error~\cite{Beasley}, and predicted vs.\ true vessel lines in the $xy$-plane~\cite{Beasley}; pressure maps, akin to deformation maps, are common in the wider palpation literature~\cite{Sanderson2020}. We omit the intermediate deformation representation and predict vessel location directly from video, the end-to-end paradigm of Bewley et al.~\cite{74}; the modular alternative of MiniTac~\cite{84} and DIGIT Pinki~\cite{80} is more interpretable, since its intermediate output can be checked against static indenter presses with accurate ground truth.

\begin{figure*}[tb]
    \begin{minipage}{1.0\linewidth}
        \centering
        \subfloat[]{%
        \includegraphics[width=0.198\textwidth]{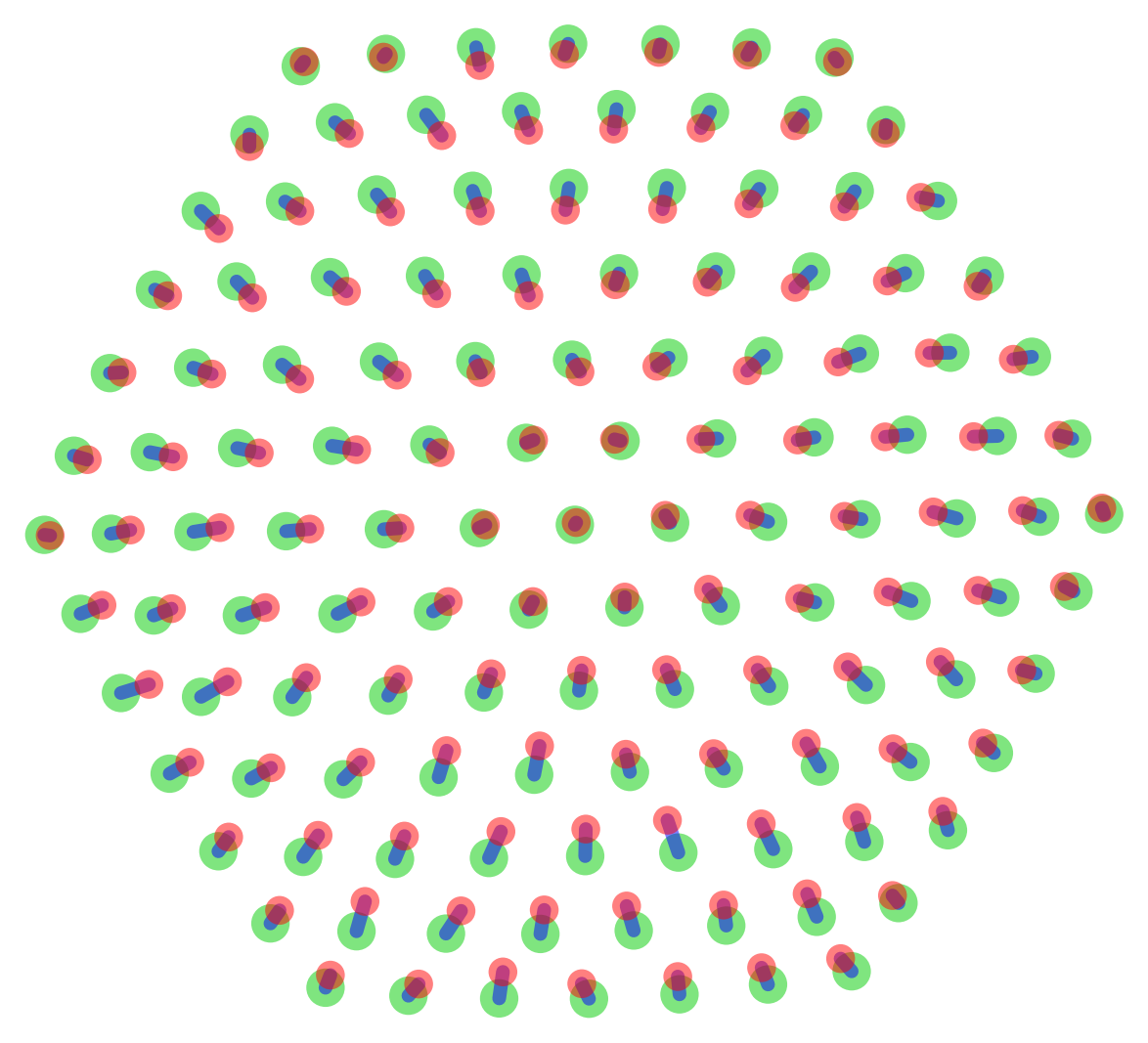}%
        }
        \hfill
        \subfloat[]{%
        \includegraphics[width=0.198\textwidth]{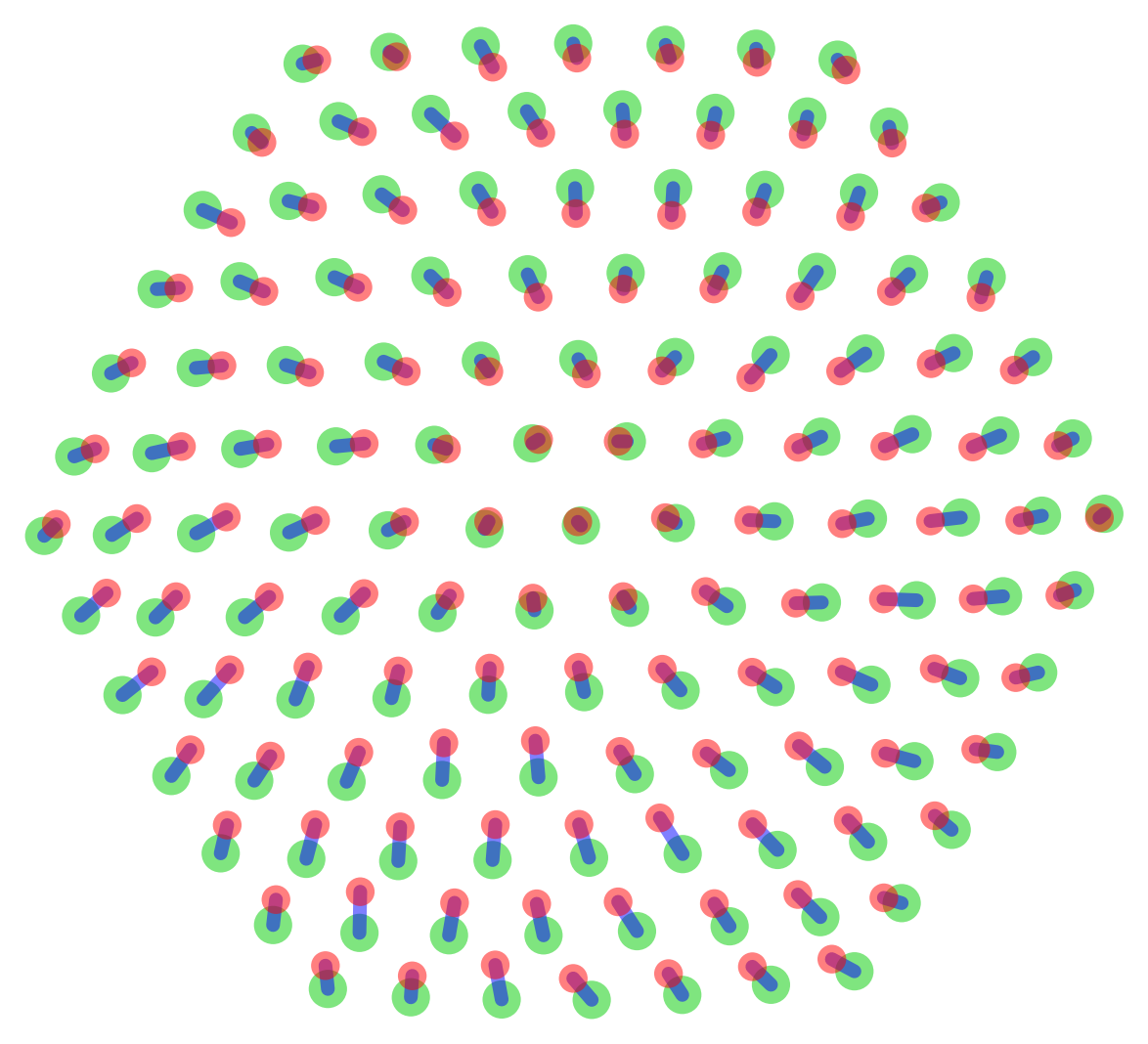}%
        }
        \hfill
        \subfloat[]{%
        \includegraphics[width=0.198\textwidth]{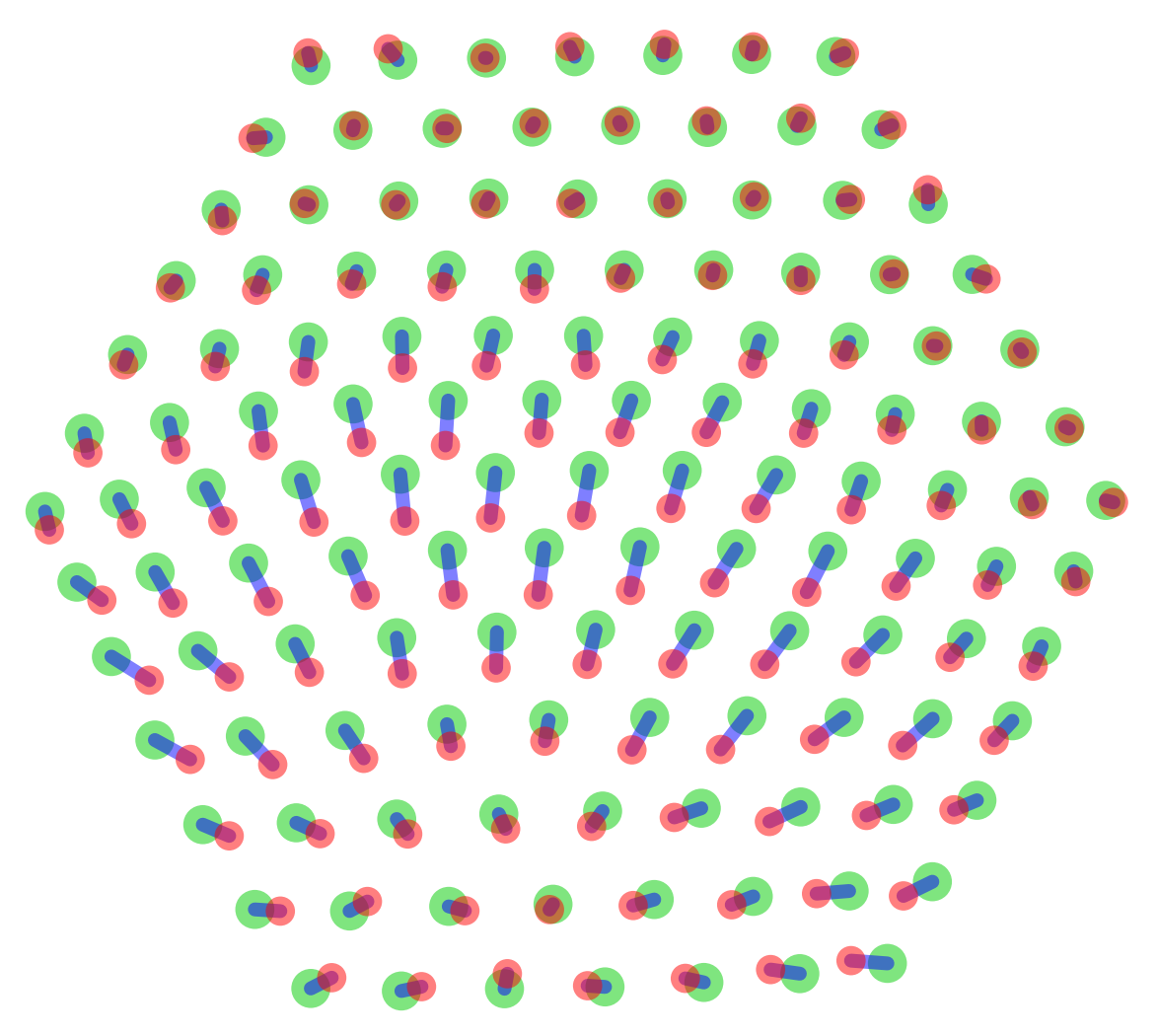}%
        }
        \hfill
        \subfloat[]{%
        \includegraphics[width=0.198\textwidth]{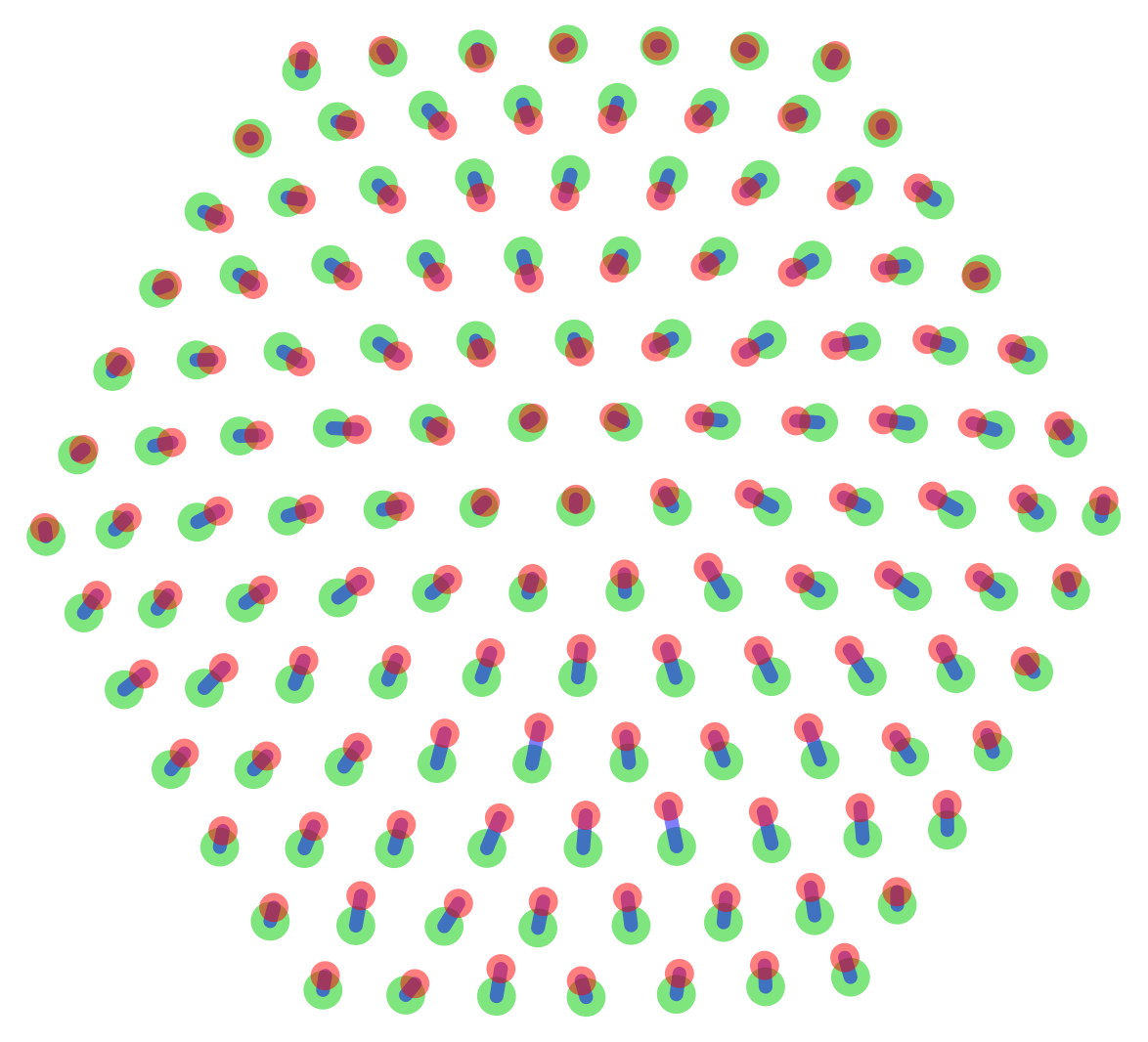}%
        }
    \end{minipage}
\caption{Alignment between simulated (red) and real (green) marker positions after domain adaptation for four canonical interactions, each shown at the point of deepest contact: (a) press, (b) twist about the $z$-axis, (c) twist about the $x$-axis, and (d) slide.}
  \label{fig:da-results}
\end{figure*}
\subsection{Meat Phantom}

\begin{figure}[tb]
    \centering
    \subfloat[]{\includegraphics[width=0.24\linewidth]{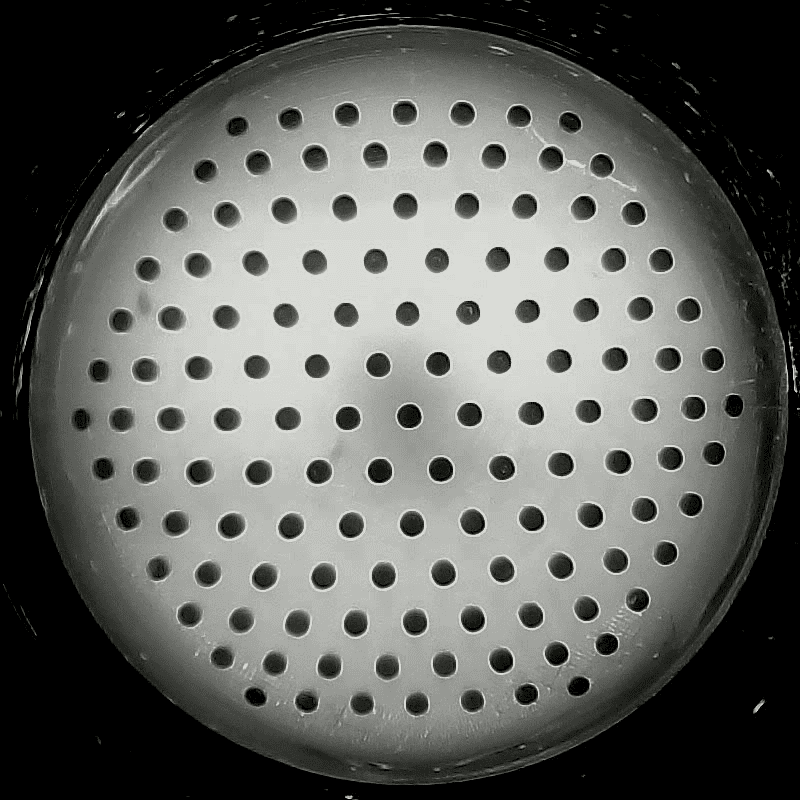}}\hspace{4pt}%
    \subfloat[]{\includegraphics[width=0.24\linewidth]{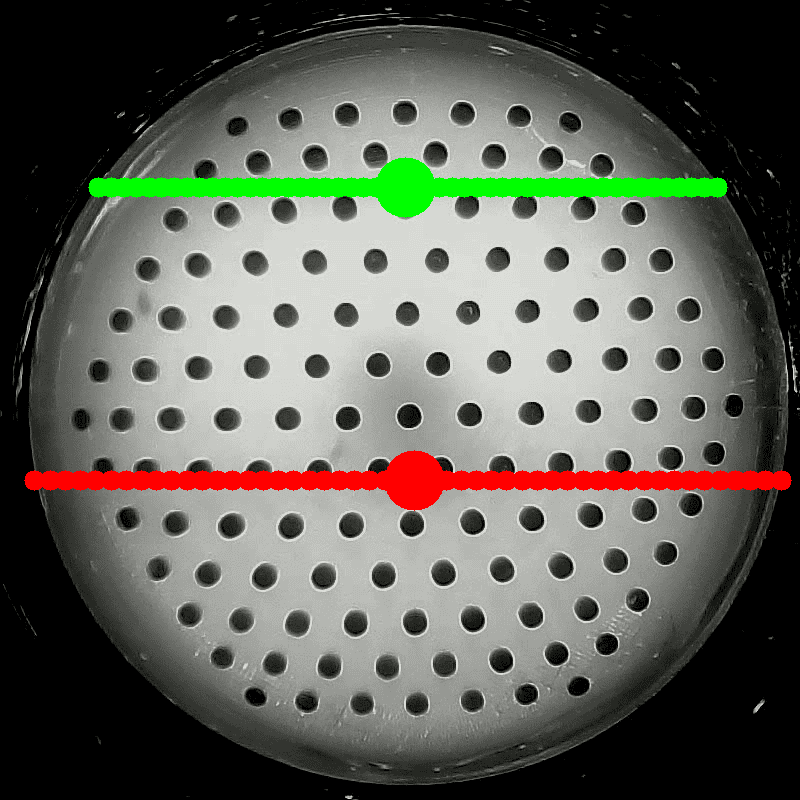}}\hspace{4pt}%
    \subfloat[]{\includegraphics[width=0.24\linewidth]{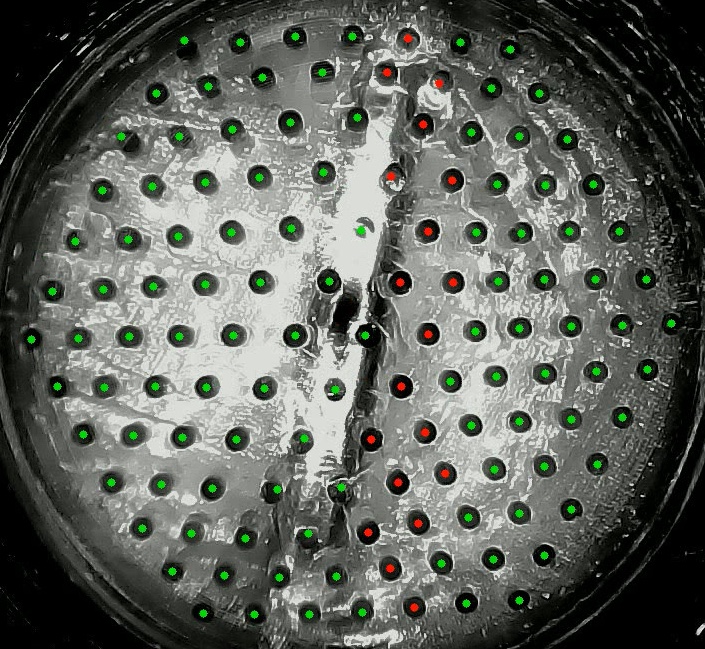}}
    \caption{(a),(b) Example Silicone video frame with 2 vessels, (a) without and (b) with annotations: large red/green circles are manually annotated points and the small-circle line is the vessel centreline from a manually specified orientation. (c) Meat ground-truth labels (red: vessel present, green: absent) from robot kinematics; note the minor misalignment with the actual metal straw.}
    \label{fig:annotation-line}
    \vspace{-0.3cm}
\end{figure}

The phantom comprised a stack of 10 raw pork medallions, each approximately 5\,mm thick. We conducted 10 single-run trials: seven with a single straw, comprising a metal straw placed beneath $n\in\{0,1,2,3,4,6\}$ medallions, corresponding to nominal depths of 0, 5, 10, 15, 20, and 30\,mm, and a silicone straw placed beneath two medallions; two trials with multiple vessels, using two and three metal straws beneath two medallions; and one control trial on the bare stack. Because the stacked tissue layers can deform and compress, these values represent nominal rather than exact vessel depths. The robot slid the sensor over the phantom (Fig.~\ref{fig:holistic-sys-arch}). The vessel surrogates were either metal straws (6\,mm OD) or silicone straws (8\,mm OD), both open-ended and air-filled at atmospheric pressure. Ground-truth straw locations were derived by combining robot kinematics with the sensor video. This automated labelling procedure simplifies data annotation but introduces a small localisation error, as illustrated in Fig.~\ref{fig:annotation-line}(c).


\section{Results and Analysis}

\subsection{Domain Adaptation}\label{sec:da-results}

This calibration of the digital twin precedes the training of any model. Over the ten BO iterations the marker misalignment on the vessel-absent slide reached at most 94.3\,px (an acquisition step towards a very soft sensor); the adopted configuration ($E = 881$\,kPa, $k_n = 9.5\times10^{4}$\,N/m), which maximises the joint objective, achieves 12.4\,px, and the misalignment on the four validation trajectories is of the same order (Fig.~\ref{fig:da-results}; mean 0.50\,mm). Simulated and real markers thus align to within a small fraction of the inter-marker spacing, the accuracy the downstream ST-GNN needs.

\subsection{Semantic Segmentation}

\begin{figure}[tb]
    \centering
    \setcounter{subfigure}{0}
    \newcommand{\prPanelH}{0.2553\linewidth}  
    \newcommand{\prPanelWa}{0.2854\linewidth} 
    \newcommand{\prPanelW}{0.2375\linewidth}  
    \includegraphics[height=0.0411\linewidth]{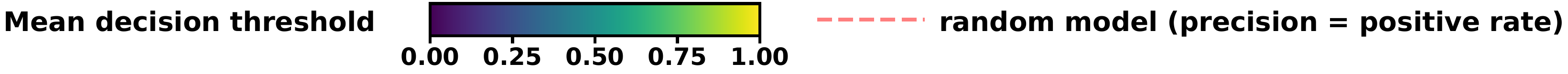}\\[3pt]
    \makebox[\linewidth][l]{%
    \includegraphics[height=\prPanelH]{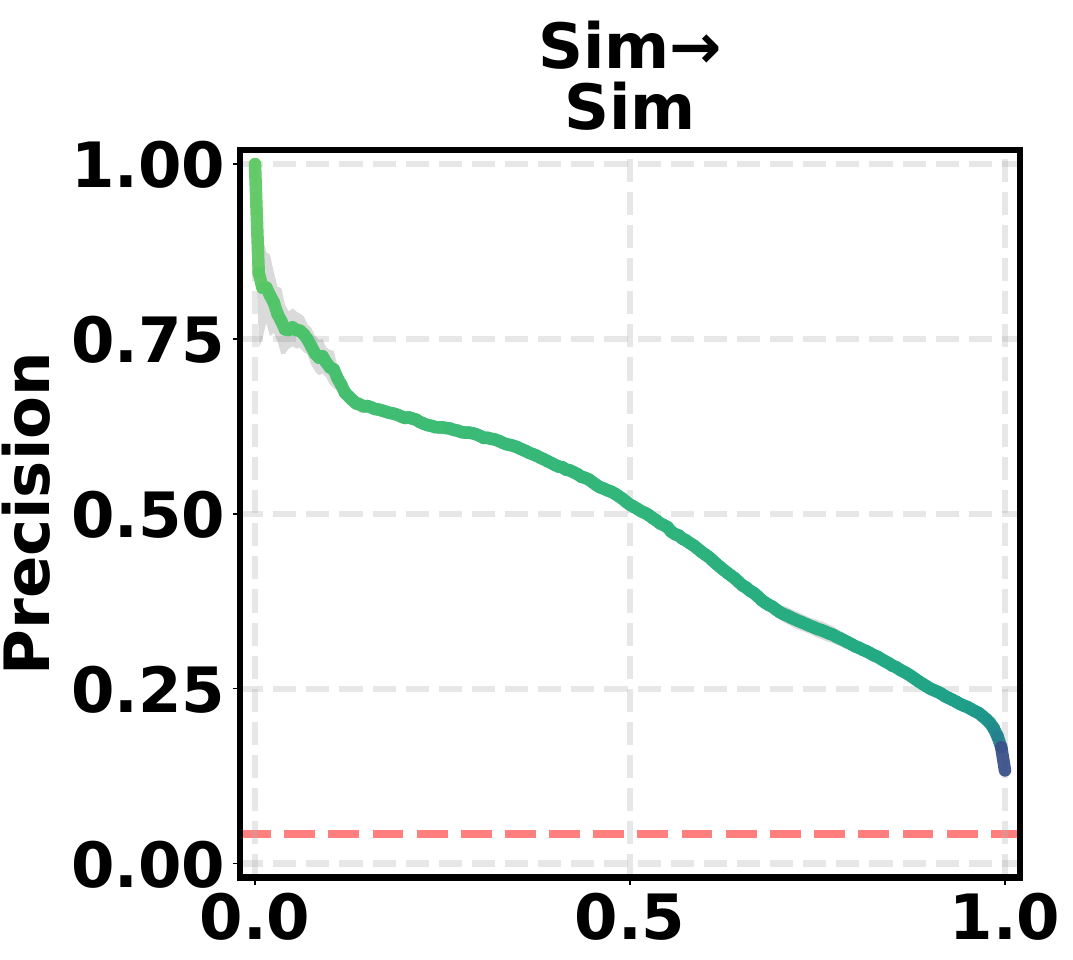}%
    \includegraphics[height=\prPanelH]{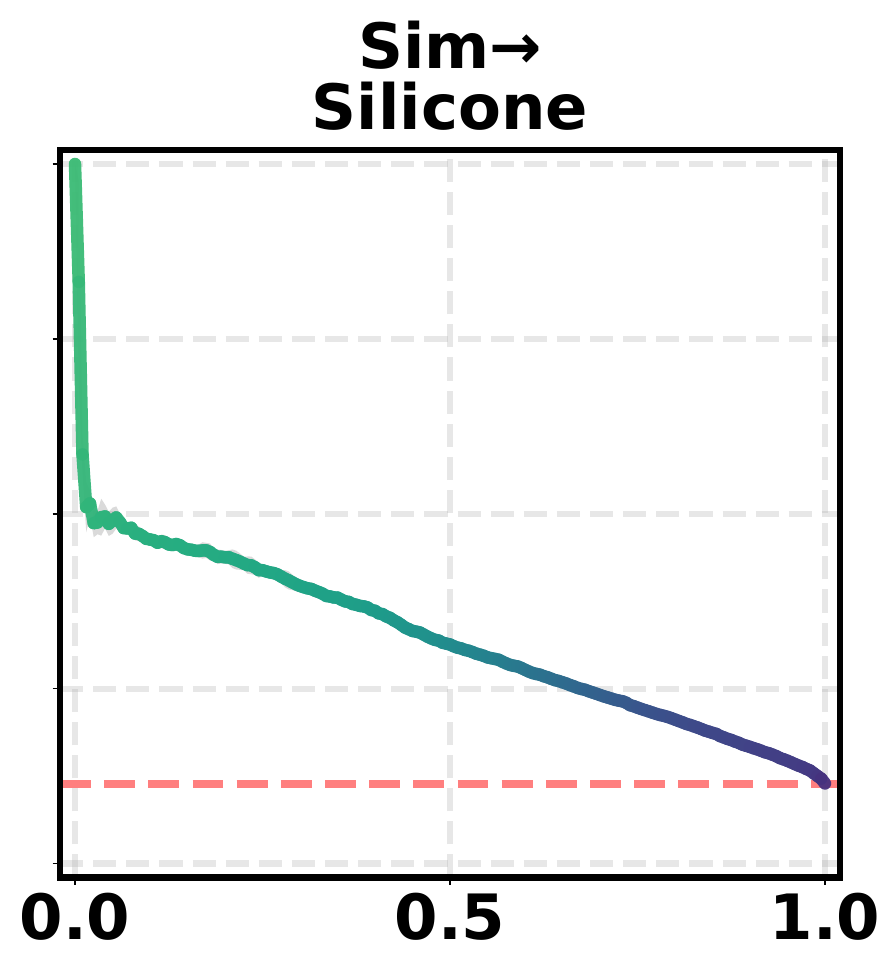}%
    \includegraphics[height=\prPanelH]{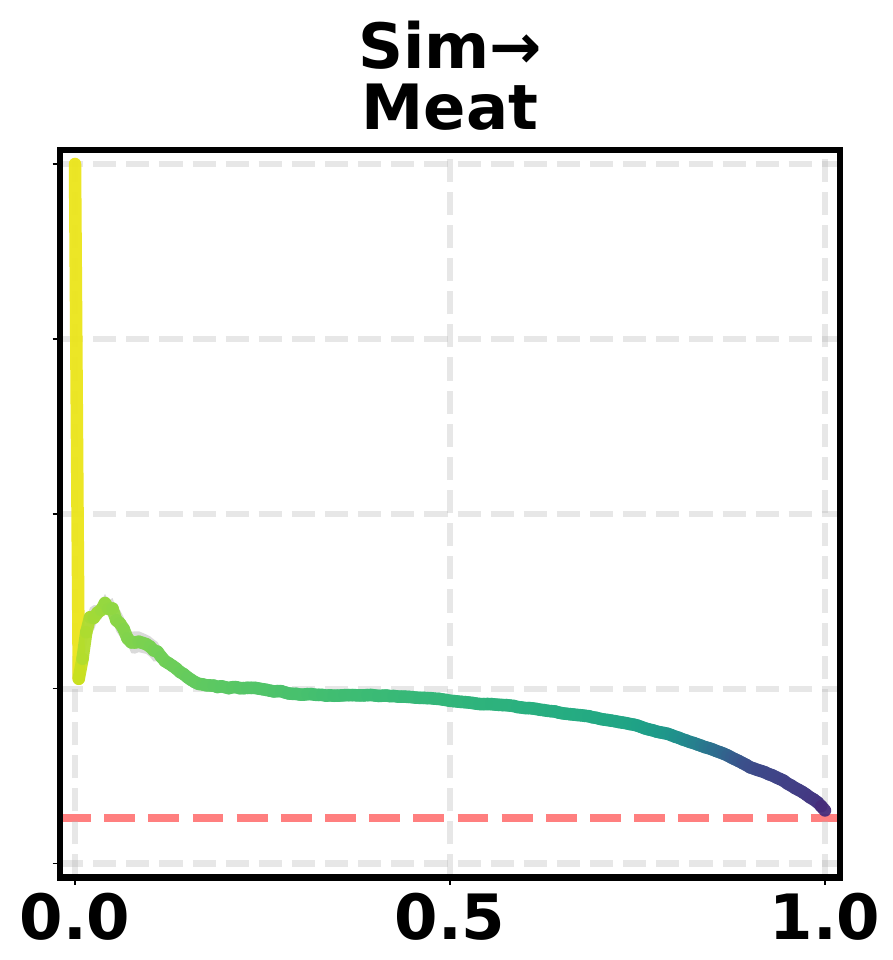}%
    \includegraphics[height=\prPanelH]{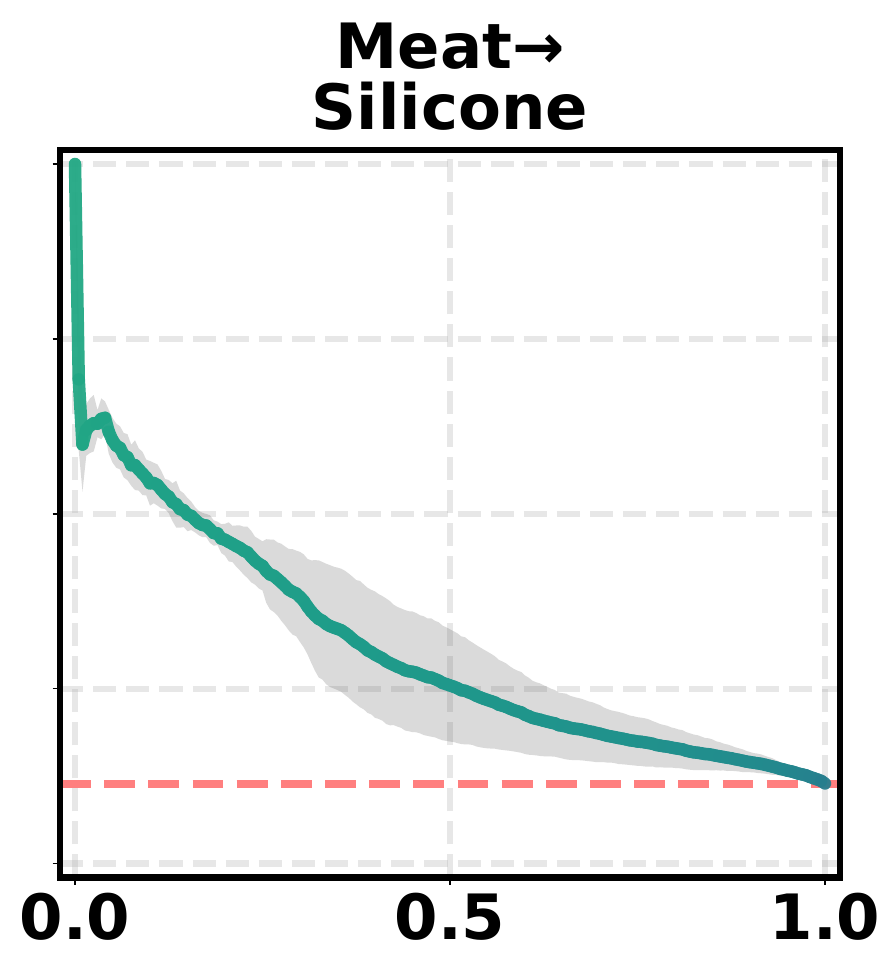}}\\[-5pt]
    \makebox[\linewidth][l]{\makebox[\dimexpr\prPanelWa+\prPanelW+\prPanelW+\prPanelW\relax][c]{\includegraphics[height=0.0177\linewidth]{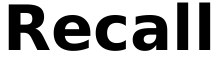}}}\\[-13pt]
    \makebox[\linewidth][l]{%
    \subfloat[]{\rule{\prPanelWa}{0pt}\label{fig:pr-curve-a}}%
    \subfloat[]{\rule{\prPanelW}{0pt}\label{fig:pr-curve-b}}%
    \subfloat[]{\rule{\prPanelW}{0pt}\label{fig:pr-curve-c}}%
    \subfloat[]{\rule{\prPanelW}{0pt}\label{fig:pr-curve-d}}}
    \caption{(a)--(d) Precision--recall curves on the test sets, over five training seeds per model: the thick line is the mean curve and the shaded band is $\pm 1$ standard deviation across seeds.}
    \label{fig:pr-curve}
    \vspace{-6mm}
\end{figure}

Fig.~\ref{fig:pr-curve} shows the threshold-free precision--recall curves over five training seeds, and the upper half of Table~\ref{tab:localisation-map} the per-marker statistics of each model's best-of-five instance, pooled once over every central frame of every test trial. Three trends stand out. First, marker-level vessel classification is hard even in domain: Sim$\rightarrow$Sim has by far the highest AP yet a foreground IoU comparable to the transfer models (and below Sim$\rightarrow$Silicone's), because the vessel occupies a thin minority of nodes and the $\sim$2\,mm marker spacing limits boundary resolution; per-node IoU is thus a diagnostic of local discrimination, and localisation is judged by the geometric aggregation of Sec.~\ref{sec:vessel-map-results}. Second, Sim$\rightarrow$Meat degrades more than Sim$\rightarrow$Silicone (lower AP and IoU), consistent with the larger gap between the simulated silicone-like phantom and layered raw meat. Third, on the shared Silicone test set Sim$\rightarrow$Silicone matches or slightly beats Meat$\rightarrow$Silicone on average (AP $0.32 \pm 0.00$ vs.\ $0.30 \pm 0.04$ over seeds, Fig.~\ref{fig:pr-curve}; foreground IoU 0.24 vs.\ 0.16 for the best instances) with far lower seed-to-seed variance, which we attribute to Sim's exact labels and much larger size (500 trajectories vs.\ 139 clips); Meat$\rightarrow$Silicone's seed spread (AP 0.26--0.35, its best instance edging Sim$\rightarrow$Silicone in Table~\ref{tab:localisation-map}) is itself a caution against comparing single-seed numbers.

\paragraph{Temporal window.} Table~\ref{tab:clip-len} shows that a single frame is clearly insufficient: the vessel signature is a temporal change in marker motion, not a static pattern. Performance rises with context up to five frames, while seven is slightly worse and markedly less stable across seeds; five frames is therefore used throughout.

\subsection{2D$\rightarrow$3D$\rightarrow$2D Semantic Segmentation Reprojection}\label{sec:vessel-map-results}

To obtain a top-view vessel map, each central-frame per-node prediction is lifted onto the sensor-tip plane through the fisheye camera model, moved into the world frame with the sensor pose from robot kinematics (Silicone, Meat) or the simulator (Sim), and dropped onto a 1\,mm/pixel grid of the phantom (the 2D$\rightarrow$3D$\rightarrow$2D projection); a pixel's score is the highest probability of any marker that landed on it, and the ground truth follows the same route from the per-node labels. Meat's ten trials are identical straight slides and share one grid (the sensor is assumed to rest undeformed on the meat); for Sim, whose published dataset lacks poses, one extra vessel-present slide was simulated with poses recorded. Maps are scored per pixel (Table~\ref{tab:localisation-map}), including the mean distance $\bar d$ from each predicted pixel to the nearest true pixel.

The decision threshold is not fixed but chosen per model as the most sensitive one that keeps map precision at or above 0.9, i.e.\ recall is maximised subject to few false alarms, because in venipuncture assistance a missed vessel merely prompts a further scan whereas a reported vessel that is not there is costly. Since the ground truth is a set of marker points rather than a filled vessel, a predicted pixel counts as correct within 3\,mm of a true pixel for this choice only; all reported statistics use exact-pixel agreement. Three models reach this operating point; Sim$\rightarrow$Meat cannot (its map precision peaks at 0.87) and is shown at its F1-optimal threshold instead. Each model is its best-of-five seed instance by AP.

\begin{figure}[tb]
\begin{minipage}{1.0\linewidth}
    \centering
    \subfloat[]{%
    \includegraphics[width=0.270\columnwidth]{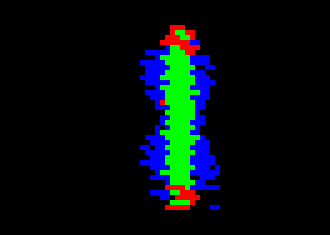}%
    \label{vm:sim}%
    }
    \hfill
    \subfloat[]{%
    \includegraphics[width=0.594\columnwidth]{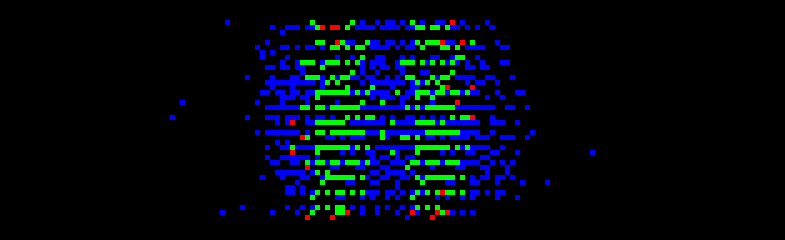}%
    \label{vm:meat}%
    }

    \vspace{3pt}

    \subfloat[]{%
    \includegraphics[width=0.441\columnwidth]{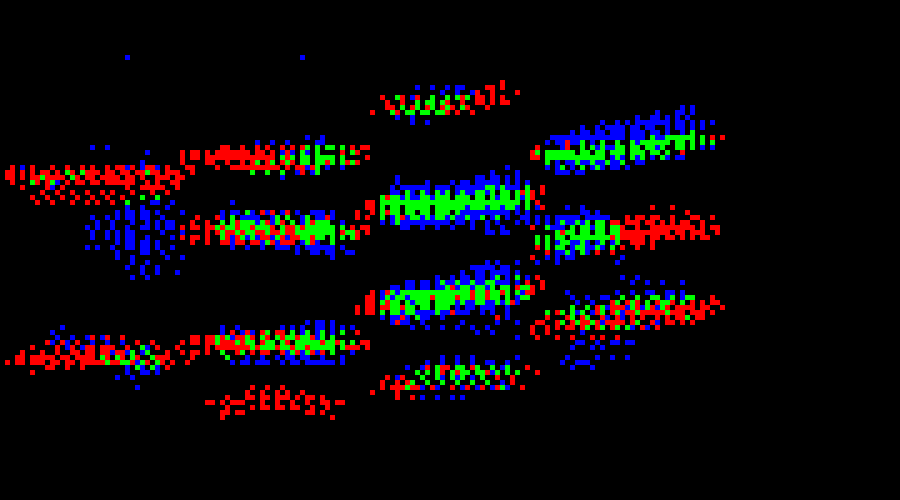}%
    \label{vm:sil}%
    }
    \hfill
    \subfloat[]{%
    \includegraphics[width=0.441\columnwidth]{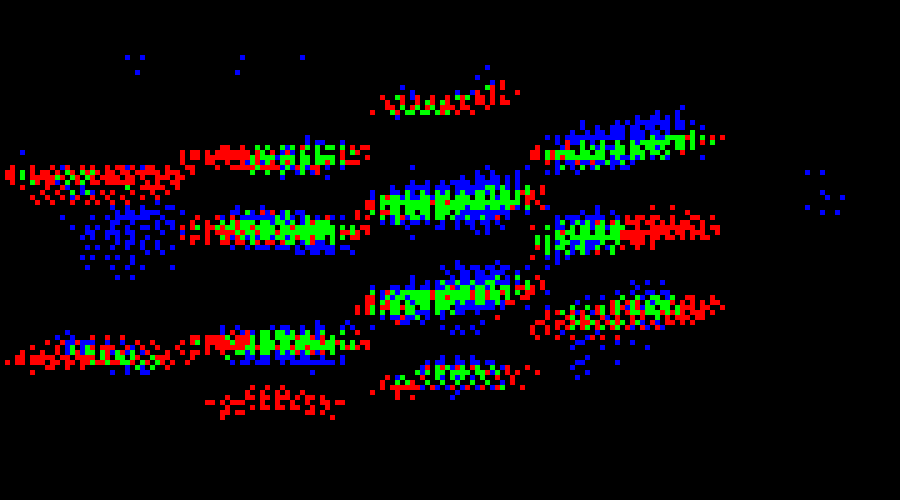}%
    \label{vm:sil2}%
    }
\end{minipage}

  \caption{Vessel localisation maps (2D$\rightarrow$3D$\rightarrow$2D geometric projection; 1$\times$1\,mm per pixel; green: TP, blue: FP, red: FN, black: TN). (a) Sim$\rightarrow$Sim, one simulated slide across the single vessel; the sensor slides horizontally across the vertical vessel. (b) Sim$\rightarrow$Meat on the trial with two metal straws beneath two medallions; the sensor slides horizontally across the vertical straws. (c) Sim$\rightarrow$Silicone and (d) Meat$\rightarrow$Silicone on the silicone phantom, which has four vessel inclusions: five vertical slides are performed, side by side, across the roughly horizontal vessels, against the video-annotation ground truth. Each model is thresholded at its own operating point (Sec.~\ref{sec:vessel-map-results}).}
  \label{fig:vessel-map}
    \vspace{-0.3cm}
\end{figure}

The maps (Fig.~\ref{fig:vessel-map}) confirm the marker-level picture. In domain (Sim$\rightarrow$Sim) predictions trace the vessel closely; on Silicone the simulation-trained and meat-trained models behave alike, clustering tightly around the vessels ($\bar d \approx 1.2$--1.3\,mm) while covering only part of their length; on Meat many predictions fall well away from the straws ($\bar d = 5.5$\,mm), although the straws themselves are recovered and the shallower ones are visibly localised. Despite the low per-node IoU, geometric aggregation therefore yields a usable localisation on the silicone phantom and in simulation, while the meat phantom exposes the limits of the current transfer.

\begin{table}[tb]
    \centering
    \footnotesize
    \setlength{\tabcolsep}{1.5pt}
    \scriptsize
    \begin{threeparttable}
    \caption{Segmentation statistics of the best-of-five instance of each model, pooled once over all its test data: per marker in video-frame space (threshold 0.5) and per pixel on the top-view map (chosen operating point). Arrows give the desirable direction; per column and within each space, \textbf{bold} marks the best value and \underline{underline} the runner-up (counts are not highlighted, as the test sets differ in size)}
    \label{tab:localisation-map}
    \begin{tabular}{lrrrrcccccccc}
    \toprule
    \textbf{Model} & TP$\uparrow$ & FP$\downarrow$ & FN$\downarrow$ & TN$\uparrow$ & MCC$\uparrow$ & F1$\uparrow$ & Prec.$\uparrow$ & Rec.$\uparrow$ & FG IoU$\uparrow$ & BG IoU$\uparrow$ & AP$\uparrow$ & $\bar d$$\downarrow$ \\
    \midrule
    \multicolumn{13}{l}{\textit{Video-frame space (per marker)}} \\
    Sim$\rightarrow$Sim & 3569 & 15169 & 38 & 66949 & \textbf{0.39} & \underline{0.32} & \underline{0.19} & \textbf{0.99} & \underline{0.19} & \textbf{0.81} & \textbf{0.50} & n/a \\
    Sim$\rightarrow$Silicone & 909 & 2070 & 818 & 11443 & \underline{0.30} & \textbf{0.39} & \textbf{0.31} & 0.53 & \textbf{0.24} & \underline{0.80} & 0.33 & n/a \\
    Sim$\rightarrow$Meat & 1493 & 7134 & 317 & 18869 & 0.29 & 0.29 & 0.17 & 0.82 & 0.17 & 0.72 & 0.23 & n/a \\
    Meat$\rightarrow$Silicone & 1600 & 8448 & 127 & 5065 & 0.20 & 0.27 & 0.16 & \underline{0.93} & 0.16 & 0.37 & \underline{0.35} & n/a \\
    \midrule
    \multicolumn{13}{l}{\textit{Top-view map space (per pixel)}} \\
    Sim$\rightarrow$Sim & 157 & 178 & 41 & 2726 & \textbf{0.58} & \textbf{0.59} & \textbf{0.47} & \underline{0.79} & \textbf{0.42} & \textbf{0.93} & \textbf{0.71} & \textbf{1.05} \\
    Sim$\rightarrow$Silicone & 592 & 750 & 752 & 15906 & 0.40 & 0.44 & \underline{0.44} & 0.44 & 0.28 & 0.91 & 0.42 & \underline{1.21} \\
    Sim$\rightarrow$Meat\tnote{$\dagger$} & 1359 & 4843 & 220 & 68938 & 0.41 & 0.35 & 0.22 & \textbf{0.86} & 0.21 & \textbf{0.93} & 0.30 & 5.49 \\
    Meat$\rightarrow$Silicone & 626 & 700 & 718 & 15956 & \underline{0.43} & \underline{0.47} & \textbf{0.47} & 0.47 & \underline{0.31} & \underline{0.92} & \underline{0.45} & 1.31 \\
    \bottomrule
    \end{tabular}
    \begin{tablenotes}[flushleft]
        \footnotesize
        \item FG/BG IoU: foreground (vessel present) / background IoU; AP: average precision (threshold-free); $\bar d$: mean L2 distance (mm) from each predicted vessel pixel to the nearest true vessel pixel. $\dagger$F1-optimal threshold (precision $\geq 0.9$ not reachable).
    \end{tablenotes}
    \end{threeparttable}
    \vspace{-0.3cm}
\end{table}

\section{Discussion}

\subsection{Comparison with Baselines}

Table~\ref{tab:comparison-2} shows that our method performs comparably to the baselines, though no direct comparison is possible: we use different phantoms, evaluate per pixel of a reprojected map rather than per trial, and localise rather than classify, so the vessel-detection works are the most suitable comparators and we do not report map accuracy (the grid is overwhelmingly vessel-free, so accuracy would be dominated by true negatives). The mean distance from a predicted vessel pixel to the nearest true one is 1.0--1.3\,mm for all models but Sim$\rightarrow$Meat, comparable to the $\sim$1\,mm mean errors of competing methods, whereas Sim$\rightarrow$Meat's 5.5\,mm is closer to the 6\,mm one of them reports under difficult conditions.

\subsection{Sliding Speed}

During the collection of Meat the robot slid at $\sim$10\,mm/s with apparently low friction (lubricated phantom, raw meat), so higher speeds (e.g., 40\,mm/s) should be feasible without damage or loss of quality, though this needs investigation.

\subsection{Limitations}
First, Silicone uses a simplified phantom with visible vessels, and both Silicone and Meat cover a limited set of trajectories; future work will target more realistic vascular phantoms and a wider range of interactions. Second, the \textit{ViTacTip}'s $\sim$2\,mm marker spacing bounds spatial resolution; finer sensing could help for very shallow or closely spaced vessels. Third, the DT needs a real-to-simulation calibration step, whose scalability and automation are future work. Fourth, the map operating point is chosen on the test data and the maps use the best-of-five seed instance, so those figures are optimistic by construction; only the seed-averaged metrics are free of this. Finally, beyond the temporal-window ablation and seed sweeps we do not isolate the contributions of the ST-GNN, domain adaptation, domain randomisation, and sliding vs.\ tapping via controlled ablations.



\section{Conclusions and Future Work}

This work presents a proof-of-concept toward trustworthy robot-assisted
sliding palpation for shallow-vessel localisation using a vision-based tactile
sensor and a calibrated digital twin. The twin is fitted to a real sliding
interaction by Bayesian optimisation and validated on four real interactions
(marker-alignment MAE 0.50\,mm), then used to generate domain-randomised
tactile sequences for training a spatio-temporal graph neural network whose
per-node predictions robot kinematics turn into an interpretable top-view
vessel map. Sim$\rightarrow$Silicone, trained without any Silicone data,
reaches an average precision of 0.32 (chance 0.11, mean over five seeds) and,
at a conservative map operating point, places predicted vessel pixels on
average 1.2\,mm from the nearest annotated vessel pixel; Sim$\rightarrow$Meat
degrades to 5.5\,mm and cannot reach that operating point, so fine
marker-level classification and cross-tissue generalisation remain
unresolved. Trustworthiness is thus meant in a limited, task-specific sense
(physical calibration, explicit sim-to-real evaluation, seed variability,
per-pixel localisation statistics and a human-verifiable output), not
clinical safety or readiness for autonomous intervention.

The digital twin is currently an offline data generator; a natural next step
is a task-specific tactile world model that predicts future tactile
observations and vessel-map updates conditioned on candidate motions, enabling
uncertainty-aware closed-loop palpation that abstains on low-confidence
predictions. Broader evaluation should cover repeated trials, held-out
calibration trajectories, varying forces and speeds, branching vessels and
non-planar surfaces, and more anatomically representative phantoms with
compliant, fluid-filled vessels, before carefully controlled human studies
under appropriate ethical approval.

\section*{AI use}

The authors used Claude (Fable, Anthropic) as an assistive tool in preparing this manuscript: to improve the conciseness and clarity of the writing, correct typographical errors, address reviewer comments, and assist in producing Fig.~\ref{fig:vitactip_working_principle}. All AI-assisted output was reviewed, verified, and edited by the authors, who take full responsibility for the content, accuracy, and integrity of the final manuscript.

%
%
\bibliographystyle{splncs04}
\bibliography{main}
\end{document}